%% file: camera_ready_supp.tex
\documentclass[10pt,twocolumn,letterpaper]{article}

\usepackage[pagenumbers]{iccv} 
\usepackage[accsupp]{axessibility}

\usepackage{multirow}
\usepackage[table]{xcolor}

\usepackage[normalem]{ulem}
\usepackage{mathtools}
\usepackage{silence}
\usepackage{multibib}
\newcites{supp}{References}

\useunder{\uline}{\ul}{}


\definecolor{iccvblue}{rgb}{0.21,0.49,0.74}
\usepackage[pagebackref,breaklinks,colorlinks,allcolors=iccvblue]{hyperref}

\def\paperID{2139}
\def\confName{ICCV}
\def\confYear{2025}

\begin{document}

\title{Factorized Learning for Temporally Grounded Video-Language Models}

\author{Wenzheng~Zeng, Difei~Gao, Mike~Zheng~Shou$^{\dagger}$, Hwee~Tou~Ng$^{\dagger}$ \\
  National University of Singapore \\ 
\tt\small wenzhengzeng@u.nus.edu, \{daniel.difei.gao, mike.zheng.shou\}@gmail.com, nght@nus.edu.sg
}
\twocolumn[{
\renewcommand\twocolumn[1][]{#1}
\maketitle
\begin{center}
    \centering
    \captionsetup{type=figure}
    \includegraphics[width=0.99\linewidth]{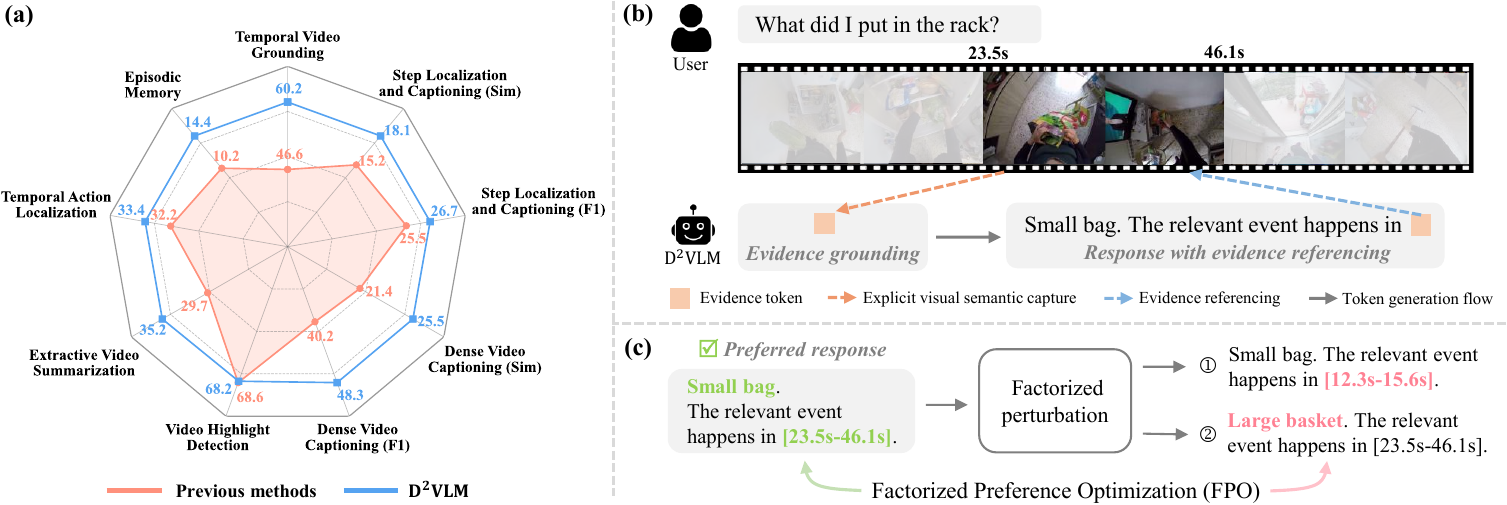}
    \vspace{-1mm}
    \captionof{figure}{(a) \textbf{Performance:} Our method outperforms SOTA methods across various tasks (here we draw the maximum performance across methods, detailed in Sec.~\ref{sec:exp}). (b) \textbf{Model}: We propose a new framework D$^2$VLM, where we decompose the generation objective into a ``grounding then answering with evidence referencing" paradigm and introduce evidence tokens to emphasize explicit event-level visual semantic capture. (c) \textbf{Training Algorithm}: We introduce Factorized Preference Optimization (FPO) that explicitly addresses both temporal grounding and textual response. A factorized data synthesis approach is also designed to support FPO.}
    \label{fig:fig1}
\end{center}
}]

\renewcommand{\thefootnote}{}
\footnotemark
\footnotetext{\vspace{-1.5em}$^{\dagger}$Corresponding authors}
\renewcommand{\thefootnote}{\arabic{footnote}}

\begin{abstract}

Recent video-language models have shown great potential for video understanding, but still struggle with accurate temporal grounding for event-level perception. We observe that two main factors in video understanding (i.e., temporal grounding and textual response) form a logical hierarchy: accurate temporal evidence grounding lays the foundation for reliable textual response. However, existing works typically handle these two tasks in a coupled manner without a clear logical structure, leading to sub-optimal objectives. We address this from a factorized learning perspective. We first propose D$^2$VLM, a framework that \underline{d}ecouples the learning of these two tasks while also emphasizing their inherent \underline{d}ependency. We adopt a ``grounding then answering with evidence referencing" paradigm and introduce evidence tokens for evidence grounding, which emphasize event-level visual semantic capture beyond the focus on timestamp representation in existing works. To further facilitate the learning of these two tasks, we introduce a novel factorized preference optimization (FPO) algorithm. Unlike standard preference optimization, FPO explicitly incorporates probabilistic temporal grounding modeling into the optimization objective, enabling preference learning for both temporal grounding and textual response. We also construct a synthetic dataset to address the lack of suitable datasets for factorized preference learning with explicit temporal grounding. Experiments on various tasks demonstrate the clear advantage of our approach. Our source code is available at \url{https://github.com/nusnlp/d2vlm}.

\end{abstract}

\section{Introduction}
\label{sec:intro} 

Recent advances in video-language models, especially those built upon large language models (video LLMs), have enabled remarkable progress in video understanding~\cite{video_llama_emnlp23, video_llava_emnlp24, video_chatgpt,timechat_cvpr24, etbench_nips24, vtgllm_aaai25}. Through their flexible video\&text-in-text-out nature, video LLMs demonstrate great potential as general-purpose solvers, unifying various tasks (e.g., temporal grounding~\cite{charades_sta_iccv17}, dense captioning~\cite{youcook2_aaai18}, and question answering~\cite{GQA_ego_cvpr24}) into one generic ``video question answering" framework.
Despite their effectiveness, existing video LLMs still struggle with accurate temporal grounding for event perception and localization~\cite{timechat_cvpr24, timechat_cvpr24, etbench_nips24, vtgllm_aaai25, trace_iclr25}, a crucial capability not only for grounding tasks themselves but also for relevant tasks that need textual answering.

We notice that for video understanding, temporal event-level grounding and textual response are two primary tasks that exhibit distinct characteristics yet maintain strong logical dependencies.
Specifically, temporal grounding focuses on precisely locating temporal events (evidence) that support answering, while textual response emphasizes accurate interpretation from the grounded evidence and generating coherent textual answers.
However, existing methods~\cite{lita_eccv24, gelm_arxiv24, etbench_nips24, timechat_cvpr24, vtimellm_cvpr24} typically handle these two tasks in a coupled manner, with two major limitations: 
(1) Various special tokens are designed for temporal grounding~\cite{lita_eccv24, gelm_arxiv24, etbench_nips24}, but their generation is mixed with text token generation without a clear logical structure, leading to coupled learning objectives.
(2) More importantly, these special tokens mainly focus on timestamp representation for precise timestamp output, lacking the explicit capture of visual semantics from grounded events. In contrast, we argue that such event-level visual semantics should not be overlooked, as it could inherently serve as crucial contexts for subsequent textual answer generation, especially under the next-token prediction paradigm.

Based on the aforementioned observations, we propose to address this from a factorized learning perspective. We first introduce a new framework D$^2$VLM that \underline{d}ecouples the learning of temporal evidence grounding and textual answering, while preserving and even strengthening their inherent \underline{d}ependency. Specifically, as shown in Fig.~\ref{fig:fig1} (b), we decompose the model response into two sequential stages: (1) pure temporal grounding that aims to localize and capture essential visual evidence for response, followed by (2) interleaved text-evidence answer generation, where both the textual answer and temporal information are produced in an evidence-referencing manner to establish consistency with the previously grounded evidence. 

Technically, we introduce evidence tokens, a special token type dedicated to temporal evidence grounding. Different from existing designs of grounding tokens that focus on its special category and timestamp representation, our evidence tokens not only aim to determine the temporal location of the grounded event, but also emphasize the capture of event-level visual semantics, which serves as crucial context for subsequent answer generation. 
During the subsequent interleaved text-evidence generation, the evidence-referencing process is achieved by generating evidence tokens that align with those from the previous grounding stage. This ensures that the output information in the final response remains consistent with the initially grounded evidence while reinforcing logical coherence across stages.
Besides providing decoupled and clearer task objectives, our design naturally fits well with the teacher-forcing autoregressive training paradigm, as subsequent textual response generation is conditioned on the correctly grounded evidence, enabling a learning shortcut for more stable training.
Our experiments demonstrate that both the designed sequence generation objective and event-level visual semantic capture are essential for performance improvement, offering valuable insights for future model design. 

To further facilitate the learning of these two tasks, we propose a novel Factorized Preference Optimization (FPO) algorithm. Unlike standard preference optimization, FPO incorporates probabilistic temporal grounding modeling into the optimization objective, enabling explicit preference learning for temporal grounding in addition to standard textual preference.  
Meanwhile, another obstacle is that existing video preference datasets do not account for such temporal grounding aspects, making it infeasible to directly apply our proposed FPO algorithm. To overcome this limitation, we construct a synthetic dataset by introducing factorized perturbations into the original preferred response sequence. These perturbations are applied at the sub-video event level, considering two main factors: temporal grounding and textual response, along with multiple possible sub-factors. This approach ensures a structured and controllable noise generation process, where the cause and type of noise are precisely known without any manual annotation. As a result, our data synthesis process is both reliable and scalable, making it possible to effectively conduct the proposed factorized preference optimization.
To the best of our knowledge, we are the first to use preference learning involving explicit temporal grounding, and moreover, in a factorized manner, as illustrated in Fig.~\ref{fig:fig1} (c). 

We conduct extensive experiments to demonstrate the superiority of our proposed framework. An intuitive comparison is shown in Fig.~\ref{fig:fig1} (a), where our 3.8B model outperforms existing SOTA methods (typically ranging from 3.8B to 13B in model size) across various tasks.

\section{Related Work}
\label{sec:related_work}

\subsection{Video LLMs}

The recent development of video LLMs has shown great potential as generic video understanding assistants~\cite{video_chatgpt, video_llama_emnlp23, video_llava_emnlp24, etbench_nips24}, primarily due to their flexible video\&text-in-text-out capability. Despite the effectiveness, existing video LLMs still struggle with accurate temporal event-level perception and localization (i.e., temporal grounding)~\cite{etbench_nips24, nextgqa_cvpr24, timechat_cvpr24, vtimellm_cvpr24,vtgllm_aaai25}, a crucial capability not only for grounding tasks but also for generating correct textual answers that are heavily based on grounding ability.

Various efforts have been made to enhance the temporal grounding ability of video LLMs~\cite{vtgllm_aaai25, lita_eccv24, gelm_arxiv24, trace_iclr25}. Besides adding targeted data for temporally sensitive fine-tuning~\cite{timechat_cvpr24, vtimellm_cvpr24, timesuite_iclr25, groundinggpt_acl24, hawkeye_arxiv24}, recent work usually designs different special tokens for precise temporal representation~\cite{vtgllm_aaai25, lita_eccv24, momentor_icml2024, etbench_nips24, grounded_videollm_arxiv24}, where some work further combines such special tokens with extra grounding decoder~\cite{gelm_arxiv24, trace_iclr25}. Despite their effectiveness, we argue that there still exist two potential limitations. First, in the generation sequence, special time tokens are interleaved with text tokens, leading to a highly coupled learning objective between temporal grounding and textual answering. Second, these special tokens are primarily designed for timestamp representations to output timestamp, lacking the ability to capture the visual semantics of grounded events.

In this work, we aim to enhance temporally grounded video LLMs from a factorized learning perspective. We first decompose the task into learning of temporal evidence grounding and textual response while also strengthening their inherent dependency. We also propose a factorized preference learning paradigm to further facilitate the two-task learning.

\subsection{Preference Optimization}
Reinforcement learning from human feedback (RLHF)~\cite{rlhf, ppo_2} or from AI feedback (RLAIF)~\cite{rlaif_1, rlaif_2} has been widely used to align LLMs' behavior with human preference, with PPO~\cite{ppo_real, ppo_1, ppo_2, ppo_3} and DPO~\cite{dpo} as common implementations. While this spirit has extended to the multi-modal domain~\cite{llava_rlhf, silkie_dpo_visual, amp_nips24}, it remains underexplored for video LLMs~\cite{dpo_video,tpo_temporal}, especially regarding explicit temporal grounding.
In this work, we introduce a factorized preference optimization (FPO) algorithm that explicitly incorporates probabilistic temporal grounding modeling into the optimization objective, enabling preference learning for both temporal grounding and textual responses. We also construct a synthetic dataset to address the lack of suitable datasets for factorized preference learning for explicit temporal grounding.

\vspace{1mm}
\section{Method}
Here we will present the proposed D$^2$VLM framework.

\subsection{Preliminary}
We begin by first illustrating the general formulation of temporally grounded video-language models, particularly in the context of video LLMs. 
Given a video $V$ and a textual question (or prompt) $Q$, the goal is to answer the question based on the given video, where the answer is typically composed of textual response together with temporal grounding information (i.e., localized temporal interval). 

For video LLMs, separate visual and text encoders will be adopted first to encode the raw inputs into a hidden space, in the format of tokens. Specifically, input video will be encoded into video tokens $F_V\in \mathbb{R} ^{\left(T\times N_v\right) \times C}$, where $T$ is the number of frames, $N_v$ is the number of frame-level visual tokens, and $C$ is the number of channels. Without loss of generality, here we set $N_v=1$ (i.e., $F_V\in \mathbb{R} ^{T\times C}$, one token per frame) for illustration purposes. Similarly, input question will be encoded into text tokens $F_Q\in \mathbb{R} ^{N_t\times C}$, where $N_t$ is the number of tokens. Then, both $F_V$ and $F_Q$ will feed into the video-aware LLM decoder, where the response will be generated in an autoregressive manner.

\begin{figure}[t]
\begin{center}
\includegraphics[width=0.45\textwidth]{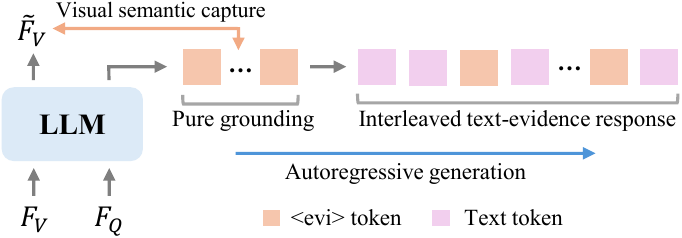}
\vspace{-5mm}
\end{center}
        \caption{Conceptual demonstration of the D$^2$VLM framework.}
\label{fig:decouple_pipeline}
\end{figure}

\begin{figure}[t]
\begin{center}
\includegraphics[width=0.38\textwidth]{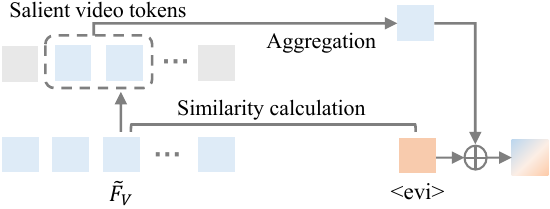}
\vspace{-5mm}
\end{center}
        \caption{The visual semantic capture process of \texttt{<evi>} token.}
\label{fig:visual_semantice_capture}
\end{figure}

\subsection{\texorpdfstring{D$^2$VLM}{D2VLM}} \label{subsec: d2vlm}
We notice that temporal event-level grounding and textual response are two primary tasks in temporally grounded video understanding tasks, where they exhibit distinct characteristics yet maintain strong logical dependencies: 
Temporal grounding focuses on precisely locating temporal events (evidence) that support answering, while textual response emphasizes accurate evidence interpretation and generating coherent textual answers. However, existing methods typically handle these two tasks in a coupled manner, with two major limitations: 
(1) Various special tokens are designed for temporal grounding, but are mixed with text token generation without a clear logical structure, leading to coupled learning objectives.
(2) More importantly, these special tokens mainly focus on timestamp representation to output timestamp, lacking the explicit capture of visual semantics from grounded events. In contrast, we argue that such visual semantics can actually serve as crucial contexts for subsequent textual answer generation, especially under the next-token prediction paradigm.

Based on this insight, we propose D$^2$VLM, where our core motivation is to decouple the tied learning objective while also preserving and even strengthening the inherent dependency between these two tasks. An overall conceptual demonstration is shown in Fig.~\ref{fig:decouple_pipeline}. Specifically, encoded video tokens $F_V$ and question textual tokens $F_Q$ are fed into the LLM decoder, which will generate response tokens autoregressively. Here, we enforce the video LLM to first conduct a pure grounding task before providing the actual answer, where the grounded events are expected to serve as essential evidence for the subsequent response. Then, the model will give the actual answer in an interleaved text-evidence manner which will be explained later.

\noindent \textbf{Evidence token for grounding and more.} Technically, we introduce evidence token \texttt{<evi>}, a special token type dedicated to temporal grounding. Unlike existing designs that mainly focus on category and timestamp representation, the proposed \texttt{<evi>} token not only identifies the temporal location of the grounded event but also emphasizes the explicit capture of event-level visual semantics. Specifically, as shown in Fig.~\ref{fig:visual_semantice_capture}, once the video LLM generates a token classified as an \texttt{<evi>} token, we first calculate the similarity between the \texttt{<evi>} token and each frame-level LLM-processed video token $\tilde{F}_V \in \mathbb{R}^{T \times C}$ (shares the same dimension as the input $F_V$). Video tokens with high similarity to the \texttt{<evi>} token are then regarded as salient tokens. We aggregate the visual semantics from these salient tokens into the \texttt{<evi>} token with simple implementations (i.e., average pool the salient tokens and then add it to the \texttt{<evi>} token). 
Such an operation explicitly endows the \texttt{<evi>} token with rich visual semantics of the corresponding grounded event, enabling it to truly serve as solid evidence for the subsequent response based on it. Experiments in Sec.~\ref{sec:exp} demonstrate that both the design of event-level modeling and visual semantic capture are crucial for performance improvement.

\noindent \textbf{Interleaved text-evi token response.}
After the pure grounding stage, the generation objective switches to answer the input question based on the grounded evidence. The answer is typically composed of textual response together with temporal grounding information (e.g., ``You capture the dog at 10s-15s and 50s-60s.''). 
Here, we formulate the response objective as an interleaved text-evidence token generation process, where \texttt{<evi>} tokens are generated again to represent the required temporal output information in a manner akin to evidence referencing, rather than using standard textual tokens to denote timestamps only.
The exact timestamps for answering can be directly obtained from the similarity calculation between the \texttt{<evi>} token and the frame-level video tokens $\tilde{F}_V$, using the frame indices of the salient video tokens (Fig.~\ref{fig:visual_semantice_capture}).
To make the model aware of such transition between tasks, we introduce another special token, \texttt{</evi>}, to indicate the end of the evidence grounding stage and encourage the model to begin generating the interleaved text-evidence answer.

Ideally, the generated \texttt{<evi>} token during this stage should also match the one in the evidence grounding phase (i.e., the final response should be consistent with the previous grounded one). We introduce an explicit constraint to enhance this consistency:
\vspace{-2mm}
\begin{equation}
    L_{cons} = \frac{1}{K}\sum_{k=1}^K \left| F_{<evi>_k}^{S_1} - F_{<evi>_k}^{S_2} \right|,
    \vspace{-1mm}
\end{equation}
where $S_1$, $S_2$ denote the pure evidence grounding stage and interleaved text-evi response stage, respectively. $k$ is the $k^{th}$ \texttt{<evi>} token in each stage. 
The $K$ \texttt{<evi>} tokens are generated in temporal order, each aligned with a distinct interval corresponding to a ground-truth event.
Such a constraint enforces strong consistency across stages, preserving logical coherence and strengthening the alignment between evidence grounding and answer generation.

\noindent \textbf{Loss function.}
We adopt the following components to supervise network learning:
\vspace{-1mm}
\begin{equation}
    L=L_{sft}+L_{gnd}+L_{cons},
\end{equation}
where $L_{sft}$ is the standard token classification loss for supervised fine-tuning~\cite{flan, flan_collection, llava_nips23}, $L_{gnd}$ is the average grounding loss for generated \texttt{<evi>} tokens across both stages, with each formulated as:
\vspace{-2mm}
\begin{equation}
    L_{gnd}^{<evi>} = \frac{1}{T}\sum_{t=1}^T BCE\left( y^t, \mathit{sim}^t \right),
    \vspace{-1mm}
\end{equation}
where BCE is the binary cross-entropy loss, $y^t$ is the frame-level ground-truth (i.e., 1/0 for foreground/background), and $sim^t$ is the normalized dot-product similarity between \texttt{<evi>} token and frame-level video token feature $\tilde{F}_V^t$.

\begin{figure*}[t]
\begin{center}
\includegraphics[width=0.9\linewidth]{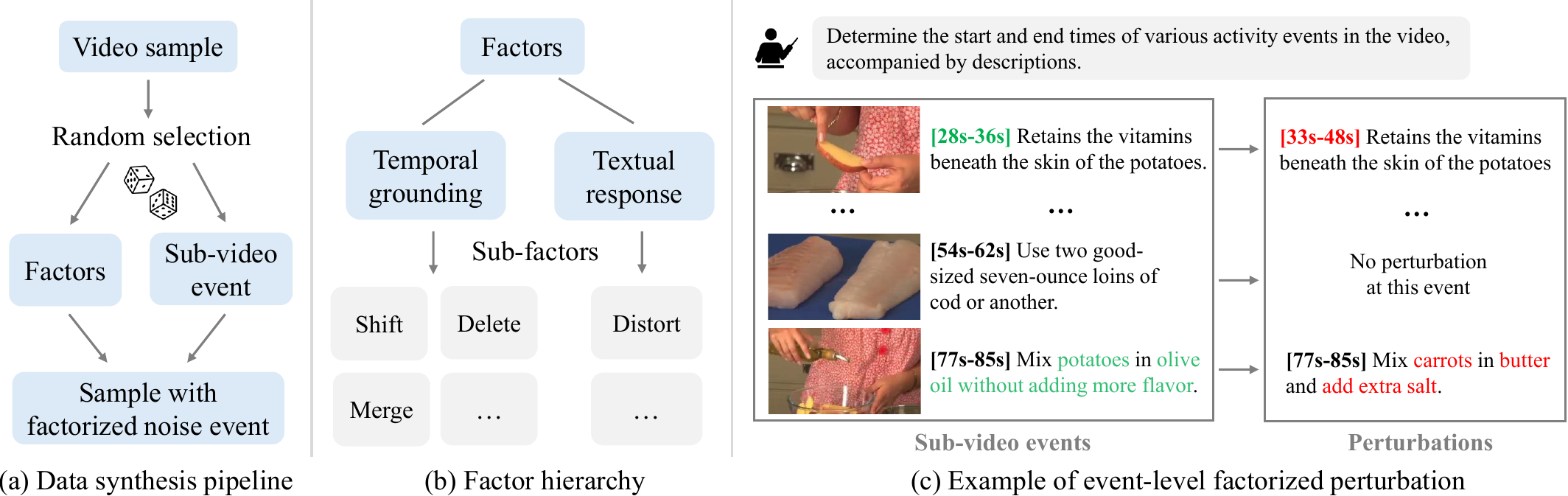}
\vspace{-4mm}
\end{center}
        \caption{The proposed data synthesis pipeline, where factorized perturbation is imposed to form the dispreferred data.}
\label{fig:data_syn}
\end{figure*}

\section{Factorized Preference Optimization} \label{sec:fpo}
To further enhance the model learning on both temporal event grounding and textual response, we introduce a novel factorized preference optimization (FPO) algorithm. Unlike standard preference optimization, FPO explicitly incorporates probabilistic temporal grounding modeling into the optimization objective, enabling explicit preference learning on both temporal grounding and textual response.
Here, we start with a macroscopic formulation that captures the essence of preference optimization and gradually elaborate on our new design.

Inspired by the existing efficient preference learning algorithm DPO~\cite{dpo}, we adopt a similar principle: encouraging the model to assign a higher probability of generating a good response than generating a bad response, which can be defined as:
\begin{equation}
L=-\left[ \log \sigma \left( \log \pi\left( R \right) -\log \pi \left( \bar{R}\right) \right) \right],  
\end{equation}
where $\pi \left( R \right)$ and $\pi \left( \bar{R} \right)$ are the joint generation probabilities of preferred and dis-preferred response, respectively. Under the autoregressive paradigm, such joint generation log-probability can be further detailed as:
\begin{equation}
    \log \pi (R)=\sum_{R_i\in R}{\log p\left( R_i\mid V,Q,R_{1:i-1} \right)},
\vspace{-1mm}
\end{equation}
where $R_i$ is the $i^{th}$ token of the whole model response $R$. $V$ and $Q$ are the input video and question, respectively.

One of our goals is to perform preference optimization not only for textual responses but also with an explicit emphasis on temporal grounding. However, achieving such multi-factor preference learning presents two key challenges: (1) The lack of suitable preference data that explicitly considers temporal grounding while maintaining a factorized structure. We will elaborate more on this issue and introduce our newly proposed synthetic dataset as a solution in Sec.~\ref{sec: data_syn}. (2) The challenge on building suitable objectives for explicit preference optimization on temporal grounding, particularly in quantifying the model's grounding probability (i.e., measuring the likelihood that the model will ground a specific temporal event).
Many existing works introduce special tokens and regression heads to enhance the grounding ability of video LLMs. Despite their effectiveness, this deviates from the standard discrete text token prediction paradigm, making it non-trivial to model grounding likelihood for preference optimization.

In this work, we notice that our designed D$^2$VLM framework can naturally model such event grounding probability, in a different and explicit way compared with the standard token prediction probability. Specifically, as grounding is achieved by frame-level feature similarity calculation between the \texttt{<evi>} token and each video frame, we directly use such similarity as the frame-level grounding probability and multiply these probabilities across frames to obtain the interval-level grounding probability. Formally, for each generated \texttt{<evi>$_k$} token during model response, its corresponding grounding probability $p_g([s_k,e_k])$ for a given time interval $\left[ s_{k,}e_k \right]$ ($s_k$, $e_k$ denote the starting and ending time index respectively) can be formulated as:
\begin{equation}
p_g([s_k,e_k])=\prod_{t=1}^T{\begin{cases}
	sim^t_k&		\mathrm{if} s_k\le t\le e_k\\
	1-sim^t_k&		\mathrm{otherwise}\\
\end{cases}}
\end{equation}
Here, $T$ is the total number of input video frames, $sim^t_k$ is the normalized dot-product similarity between $<$evi$>_k$ and frame-level video token feature $\tilde{F}_V^t$. Based on such event grounding modeling, the log-probability formulation within our proposed factorized preference optimization is given as:
\begin{equation}
\begin{aligned}
	\log \pi (R) &=\sum_{R_i\in R}{\log p\left( R_i\mid V,Q,R_{1:i-1} \right)}\\
	&+ \sum_{\mathclap{R_k = <evi>}}  
	{\log p_g\left( \left[ s_k,e_k \right] \mid V,Q,R_{1:k-1} \right)},
\end{aligned}
\end{equation}
where the first term corresponds to the standard textual token prediction modeling, while the second term represents the proposed explicit modeling on temporal event grounding. By substituting Equation (7) into Equation (4), we obtain the final formulation of our Factorized Preference Optimization (FPO). Next, we explain the process of constructing multi-factor preferred ($R$) and dis-preferred ($\bar{R}$) data pairs, which serve as the training data for FPO.

\section{Factorized Preference Data Synthesis} \label{sec: data_syn}

As previously mentioned, one key challenge is the lack of suitable preference data that captures multi-factor preferences—not only for textual responses but also for explicit temporal grounding.
Particularly, in the video domain, preference data is typically constructed by providing a video LLM with either complete or degraded video inputs (e.g., blurred or cropped images/videos) and treating its responses as preferred or dispreferred, respectively~\cite{tpo_temporal}. The underlying assumption is that the model will produce inferior responses when given degraded or incomplete visual inputs. 
However, existing works using this approach often face some major issues: (1) The reliability of the generated preference data is relatively less controlled. Specifically, the generated preferred sample may not actually be the preferred one, and the generated dispreferred data can sometimes be as good as or even better than the preferred data~\cite{tpo_temporal}. As a result, post-filtering is usually required, which involves additional models and ad-hoc human interventions, increasing costs but still cannot fully ensure quality. (2) This lack of control also makes it difficult to properly factorize the generated dispreferred data, including identifying the source or the reason why it is considered dispreferred. (3) Last but not least, existing video preference data mainly focuses on generating responses in pure textual format (e.g., video caption)~\cite{dpo_video, tpo_temporal}, while generating preference response that explicitly contains temporal grounding information (e.g., temporal event position) remains underexplored. Overall, these issues make factorized preference optimization difficult to achieve.

To address the aforementioned limitations, we propose an automatic factorized preference data synthesis framework, which introduces factorized perturbations on the original response at the sub-video event level. As shown in Fig.~\ref{fig:data_syn} (a), the overall process consists of the following steps:

\noindent \textbf{Random selection of factors and sub-video events.}
For each given sample (video-question-answer pair), we first randomly select a factor that determines the exact type of the perturbation. As shown in Fig.~\ref{fig:data_syn} (b), the factors are broadly categorized into two types and each can be further divided into multiple sub-types:
(1) Temporal grounding: This type of perturbation aims to reveal/simulate model response inaccuracies related to temporal event grounding. It can be further divided into sub-types, including temporal localization shift, randomly adding/deleting grounded events (corresponding to false positives and missed detections), and merging multiple events into one (corresponding to a lack of fine-grained distinction in event boundaries).
(2) Textual response: This type of perturbation modifies the semantic correctness of the textual response. It includes sub-types such as distorting key information, which disrupts critical content, and repeating responses, a common failure mode observed in video LLMs.

Next, we randomly select some sub-video events to apply factorized perturbation. A sub-video event refers to a semantically meaningful segment of the video that can be determined based on the given sample (e.g., its question or ground-truth answer). For instance, in a dense video captioning task, an event naturally corresponds to a specific captioned interval, while in an action localization task, an event can align with a single action instance. Without loss of generality, we take the dense video caption task as an illustration example (as shown in Fig.~\ref{fig:data_syn} (c)), as this task emphasizes both temporal grounding and textual description. Overall, such an event-level design aligns naturally with the inherent video structure, ensuring that perturbations are introduced in a way that respects the video's temporal and semantic coherence. Moreover, it enables fine-grained perturbation, allowing controlled modifications at the event level rather than applying coarse, less precisely controlled noise to the entire response sequence.

\noindent \textbf{Injecting factorized perturbations.} 
Once the factors and sub-video events are determined, localized perturbation is introduced based on the selected factors and sub-video events. If a temporal grounding factor is selected, we modify the timestamps of the selected events based on the determined sub-factor (e.g., shift the interval, delete the interval, etc.). For the factor of textual response, for each selected event, we prompt off-the-shelf LLM (e.g., Qwen~\cite{qwen2,qwen2.5}) to generate a distorted response based on the original correct event-level response. If the ``repeat" sub-factor is selected, we can easily achieve this by setting the response of the current event to its previous one to simulate such a response repeating phenomenon. 
A synthesized example is shown in Fig.~\ref{fig:data_syn} (c). After the perturbation synthesis, we combine the newly obtained noisy response with the original one to form the preference sample pair for factorized preference optimization. In practice, our preference data is synthesized based on the E.T. Instruct 164K dataset~\cite{etbench_nips24}.

\noindent \textbf{Discussion.} Compared to existing preference data synthesis approaches, our framework ensures that the generated dispreferred response contains specific, well-factorized noise rather than arbitrary, unpredictable degradation, and thus we can better identify the noise source without manual costs. To our knowledge, we are also the first to emphasize explicit temporal grounding in preference data, and moreover, in a multi-factorized manner. The overall design makes our factorized preference optimization in Sec.~\ref{sec:fpo} feasible.

\section{Experiments} \label{sec:exp}

\subsection{Experimental Setup}

\noindent \textbf{Implementation details.}
Following the practice of~\cite{etbench_nips24}, we adopt the pre-trained ViT-G/14 from EVA-CLIP~\cite{eva, eva_clip}, followed by a Q-Former-like feature compressor~\cite{etbench_nips24,blip2,instruct_blip_nips23}, and use Phi-3-Mini-3.8B~\cite{phi3_arxiv24} as the base LLM. Also following~\cite{etbench_nips24}, our model is initialized from the stage-2 model of~\cite{etbench_nips24} and~\cite{llama_vid}, and we also only fine-tune our model on the E.T. Instruct 164K dataset~\cite{etbench_nips24} using LoRA~\cite{lora}. 
We conduct training on 4 $\times$ NVIDIA H100 GPUs within 1 day. More details are provided in the supplementary material.

\noindent \textbf{Evaluation tasks and datasets.}
We evaluate our method across various datasets and tasks. (1) E.T. Bench Grounding~\cite{etbench_nips24}. It consists of 5 different grounding-related tasks (e.g., moment retrieval, action localization, summarization/highlight detection, etc.) for comprehensive evaluation. (2) E.T. Bench Dense Captioning~\cite{etbench_nips24}. It is composed of tasks like dense video caption and step localization, where both key event grounding and textual description are required. Besides, we also conduct experiments on two widely used individual datasets, (3) Charades-STA~\cite{charades_sta_iccv17} and (4) YouCook2~\cite{youcook2_aaai18}, for independent evaluation on moment retrieval and dense video captioning, respectively.

\input{ICCV2025-Author-Kit-Feb/tab/quan}

\subsection{Main Benchmarking Results}

\noindent \textbf{E.T. Bench grounding.} The results are shown in the left part of Tab. \ref{tab:quan}, where we compare our method with the recently published SOTA methods. It can be observed that the proposed D$^2$VLM significantly outperforms recently published SOTA methods across various tasks, achieving an average F1 score improvement of at least 7.0\%.
Meanwhile, our model is relatively lightweight (3.8B parameters) compared to most state-of-the-art methods (ranging from 3.8B to 13B, with the majority being 7B parameters). This further demonstrates the superiority of our proposed method.

\noindent \textbf{E.T. Bench dense captioning.} The performance comparison is shown in the right part of Tab.~\ref{tab:quan}. It can be seen that our method also shows noticeable superiority in dense captioning-related tasks, where it outperforms recent SOTA methods in both temporal event grounding (at least 5.3\% on average F1 score) and textual description (at least 3.6\% on average Sim score). This essentially demonstrates the generalizability of our method across various scenarios.

\noindent \textbf{Charades-STA.} From Tab.~\ref{tab: charades}, it can be seen that our method still outperforms recent SOTA methods by large margins. Specifically, our 3.8B model outperforms counterpart~\cite{etbench_nips24} with the same base model by 4.4\% on R@1(IoU=0.5), and outperforms methods published in 2025 with 7B model size~\cite{vtgllm_aaai25,trace_iclr25,timechat_cvpr24} by at least 1.6\% on R@1(IoU=0.5).  This further demonstrates the superiority of our method in both effectiveness and efficiency.

\noindent \textbf{YouCook2.} As shown in Tab.~\ref{tab: youcook2}, our method achieves state-of-the-art performance compared to recent SOTA approaches. Specifically, it outperforms them by at least 4\% in temporal event grounding (F1) while also demonstrating superior textual description capacity, with gains of at least 2.5\% on CIDEr and 1.0\% on SODA\_c. This further highlights the superiority of our approach in both temporal grounding and text generation.

\subsection{Ablation Studies}

Here we conduct ablation studies on E.T. Bench grounding and E.T. Bench Dense Captioning datasets to analyze the effect of the proposed components, in both temporal event-level grounding and textual description. For each dataset (i.e., grounding, dense captioning), we report the average performance across multiple sub-tasks. In the following, we first analyze the component effect of the newly designed generation objective and evidence token, where we do not take the proposed FPO into account. Finally, we analyze the effect of the proposed FPO.

\input{ICCV2025-Author-Kit-Feb/tab/ab1}

\input{ICCV2025-Author-Kit-Feb/tab/ab2}

\noindent \textbf{Generation objective of D$^2$VLM.}
Here, we analyze the effectiveness of our newly designed generation objective within D$^2$VLM.
From Tab. \ref{tab:ab1}, it can be seen that: (1) Row-1 v.s. Row-2: Decompose the coupled grounding-textual response objective into grounding (by our designed \texttt{<evi>} token) then answer from grounded event can substantially improve the performance in both grounding (i.e., 7.7\% averaged F1 on grounding tasks and 8.8\% averaged F1 on dense captioning tasks) and textual description (i.e., 4.7\% averaged sentence similarity on dense captioning tasks). (2) Row-2 v.s. Row-3: When we shift the answering stage from pure textual token generation into interleaved text-evi token generation, the performance can be further enhanced by large margins (at least 6.7\% on grounding and 3.8\% on textual description). The reason is that under such cases, the overall generation objective can be interpreted as grounding then answer with evidence referencing, where such referencing strengthens the relationship between event grounding and textual answering. (3) Row-3 v.s. Row-4: The introduced consistency constraint further boosts the performance by a noticeable margin, as it explicitly enhances the consistency between the two stages. Overall, the results validate the effectiveness of our design.

\noindent \textbf{Design of evidence token.}
Here we analyze the impact of the core components within the \texttt{<evi>} token design (i.e., the event-level modeling and the explicit visual semantic capture). The results are shown in Tab. \ref{tab:ab2}, where w/o event-level modeling means each \texttt{<evi>} token only captures the information of individual frame (a common design in existing works~\cite{etbench_nips24, lita_eccv24, momentor_icml2024}), rather than an interval-level event (e.g., 10s-15s). w/o visual semantic capture indicates visual semantics from the video tokens is not explicitly integrated into the \texttt{<evi>} token through the operation illustrated in Sec.~\ref{subsec: d2vlm} (also a common design in most of the existing works~\cite{lita_eccv24, momentor_icml2024, gelm_arxiv24}). From Tab. \ref{tab:ab2}, it can be summarized that: (1) The performance drops significantly without event-level modeling. We argue that this is because compared to individual timestamp-level modeling, event-level modeling more naturally aligns with the characteristics of event-level evidence grounding and can capture more robust event-level semantics in a global perspective. (2) Explicit visual semantic capture is also crucial for enhancing model effectiveness, as it enables \texttt{<evi>} tokens to truly serve as solid context for subsequent answer generation, which also fits well with the autoregressive generation process. Notably, compared with grounding-only tasks, such an effect is more pronounced on dense captioning tasks, as textual description generation can truly benefit from the captured event-level visual semantics.

\noindent \textbf{Effect of factorized preference optimization.}
Tab. \ref{tab:ab3} demonstrates that the proposed FPO algorithm consistently enhances performance on both temporal grounding and captioning-related tasks, with a more significant improvement in temporal grounding. This indeed validates the effectiveness of our factorized optimization objective design.

\section{Conclusion and Limitations}

In this work, we uncover some key limitations of existing temporal grounded video-language models: (1) the use of coupled learning objective for temporal grounding and textual response generation, and (2) the overlooking of inherent dependency between these two tasks, especially the event-level visual semantics from the intermediate grounding result. We tackle this from a factorized learning perspective, where we propose D$^2$VLM that decomposes the generation objective into a “grounding then answering with evidence referencing” paradigm, and introduce evidence tokens to emphasize explicit visual semantic capture beyond timestamp representation. Furthermore, we design a factorized preference learning algorithm, coupled with a factorized synthetic dataset, to enhance the learning of both tasks. Extensive experiments demonstrate the clear superiority of our approach across various scenarios. 

Despite achieving new state-of-the-art results, our approach still has room for improvement (e.g., F1 scores of only 14.4\% on the episodic memory task and 26.4\% on the YouCook2 dense video captioning task). Besides, while our factorized data generation offers good factorized controllability and directly contributes to factorized optimization, it focuses only on generating negative (dis-preferred) samples. However, generating more diverse positive samples to further enrich the preference data is also meaningful, and we leave this for future work.

\section*{Acknowledgment}
This research is supported by the National Research Foundation, Singapore under its AI Singapore Programme (AISG Award No: AISG3-RP-2022-030).
We would like to acknowledge that computational work involved in this research work is partially supported by NUS IT’s Research Computing group under grant number NUSREC-HPC-00001.

{
    \small
    \bibliographystyle{ieeenat_fullname}
    \bibliography{main}
}

\clearpage
\appendix

\maketitlesupplementary

\setcounter{figure}{0}
\setcounter{table}{0}
\setcounter{footnote}{0}

\section{More Implementation Details}

Here we provide more implementation details in addition to the main paper. 

Following existing practice~\citesupp{llama_vid_supp,etbench_nips24_supp}, we adopt 1 FPS frame sampling for both training and testing. Frames are resized to $224 \times 224$ before being fed into the network. 

To determine the salient tokens for grounding and explicit event-level visual semantic capture, we treat frames within the ground-truth interval as salient during training. During inference, if the feature similarity between a frame-level video token and \texttt{<evi>} exceeds 60\% of the maximum similarity between the current \texttt{<evi>} token and all frame tokens, the corresponding frame-level video token is considered salient and included. Sec.~\ref{exp:threshold} demonstrates the performance robustness to threshold variation. 

Before similarity calculation, the \texttt{<evi>} token is first projected through a 2-layer MLP. This projection helps to distinguish its two functional roles: serving as a generation token during autoregressive decoding (via a standard LM classification head), and acting as a query token for similarity-based grounding and visual semantic aggregation. This design facilitates the joint learning of these related but functionally distinct tasks.

For the performance comparison among different methods, most reported numbers are directly taken from the original papers, except for Qwen2.5-VL~\citesupp{qwen2.5vl_supp} on E.T. Bench~\citesupp{etbench_nips24_supp}, which we re-implemented due to the absence of official results. We found that the performance is highly sensitive to the prompt and pixel configurations, which aligns with findings discussed in the context of video temporal grounding on GitHub\footnote{\url{https://github.com/QwenLM/Qwen2.5-VL/issues/837}}.
We combine the official cookbook from Qwen2.5-VL and the practice from lmms-eval\footnote{\url{https://github.com/EvolvingLMMs-Lab/lmms-eval}}, resulting in considerably higher performance compared with directly using the official cookbook which may not be tailored for benchmarking purposes. We report the highest performance of Qwen2.5-VL that we were able to achieve in our paper.

\begin{figure*}[ht]
\centering
\includegraphics[width=0.79\textwidth]{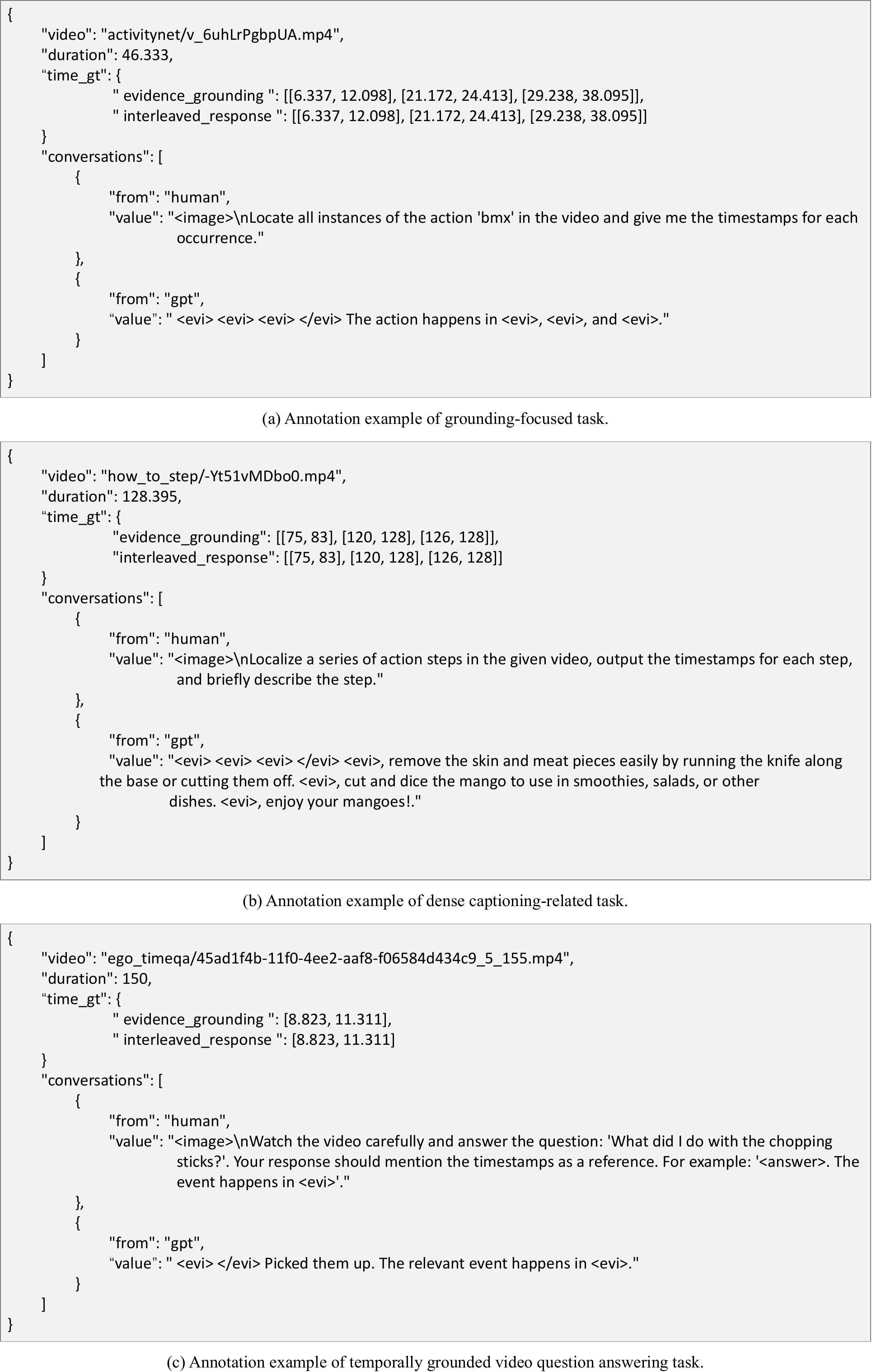}
\caption{Qualitative examples for grounding-focused task, dense captioning-related task, and temporally grounded video question answering task.}
\label{fig:anno}
\end{figure*}

\section{Data Annotation Formats}
Here, we also provide the annotation formats for model training, which offer an intuitive and clear understanding of D$^2$VLM's generation objective. Based on the input-output format, we categorize different tasks into three main types: (1) Grounding-focused task (e.g., temporal video grounding, action localization, etc.), which can involve single-event grounding and multi-event grounding. (2) Dense captioning-related task, which requires grounding multiple events throughout the entire video while also providing a textual description for each grounded event. (3) temporally grounded video question answering, which involves answering the user's open-ended questions while also providing the temporal position of the answer (evidence). 

The examples are shown in Fig.~\ref{fig:anno}. 
Here, we provide the definitions of some important keys in the annotation files. The ``conversations" key is the main component, which consists of two sub-parts: ``from human" and ``from gpt".  
The value corresponding to the ``from human" part represents the input prompt, which mainly includes the video (represented here as a place-holder \texttt{<image>}, but will be actually replaced by video frames) and the user question (instruction).  
The second part, ``from gpt", represents the desired model response sequence, which typically consists of two stages: the pure evidence grounding stage and the interleaved text-evidence token generation stage. These two stages are separated by the \texttt{</evi>} token, which the model should also generate to indicate the end of the evidence grounding stage and the beginning of the interleaved response.
Another important key is ``time\_gt," which indicates the ground-truth temporal event position. This is used to supervise the similarity calculation between the \texttt{<evi>} token and frame-level tokens, as mentioned in this paper. Here, the ground-truth annotations for the evidence grounding stage and the interleaved response stage are the same, based on the natural assumption that the grounded evidence should be consistent with the answer.

\begin{table}[t]
\centering
\footnotesize
\begin{tabular}{lccc} \toprule
Method            & Year       & TEM (Rec)     & GVQ (Rec)    \\ \midrule
Video-ChatGPT-7B~\citesupp{video_chatgpt_supp}  & ACL'24     & 15.9          & 0.0            \\
Video-LLaVA-7B~\citesupp{video_llava_emnlp24_supp}    & EMNLP'24   & 7.5           & 0.1          \\
LLaMA-VID-7B~\citesupp{llama_vid_supp}      & ECCV'24    & 7.0           & 0.9          \\
Video-LLaMA-2-7B~\citesupp{video_llama_2_supp}  & arXiv'24   & 0.0           & 0.1          \\
PLLaVA~\citesupp{pllava_supp}            & arXiv'24   & 4.1           & 1.2          \\
VTimeLLM-7B~\citesupp{vtimellm_cvpr24_supp}       & CVPR'24    & 6.8           & 1.9          \\
VTG-LLM-7B~\citesupp{vtgllm_aaai25_supp}        & AAAI'25    & 8.9           & 1.4          \\
TimeChat-7B~\citesupp{timechat_cvpr24_supp}       & CVPR'24    & 18.0          & 1.5          \\
LITA-13B~\citesupp{lita_eccv24_supp}          & ECCV'24    & 16.0          & 2.2          \\
E.T. Chat-3.8B~\citesupp{etbench_nips24_supp}     & NeurIPS'24 & 16.5          & 3.7          \\
\rowcolor[HTML]{ECF4FF}
D$^2$VLM-3.8B (Ours) & ICCV'25       & \textbf{29.2} & \textbf{7.1} \\ \bottomrule
\end{tabular}
\caption{Performance comparison on TEM (Temporal Event Matching) and GVQ (Grounded Video Question answering).}
\label{tab: complex}
\vspace{-5mm}
\end{table}

\begin{table}[t]
\centering
\footnotesize
\begin{tabular}{lcc}
  \toprule
  Method & MVBench & \begin{tabular}[c]{@{}c@{}}Video-MME \\ (w/o subs)\end{tabular} \\ \midrule
  Video-LLaVA-7B~\citesupp{video_llava_emnlp24_supp} & 43.0 & 39.9 \\
  E.T. Chat-3.8B~\citesupp{etbench_nips24_supp}   & 36.4 & 34.5 \\
\rowcolor[HTML]{ECF4FF}
  D$^2$VLM-3.8B (Ours)      & \textbf{43.9} & \textbf{43.9} \\ \bottomrule
  \end{tabular}
  
  \caption{Performance comparison on general video-question-answering benchmarks.}
  \label{tab:general_video}
\end{table}

\begin{table}[t]
\centering
\footnotesize
\begin{tabular}{c|c|cc}
\toprule
\multirow{2}{*}{Threshold} & Grounding  & \multicolumn{2}{c}{\begin{tabular}[c]{@{}c@{}}Dense Captioning\end{tabular}} \\
                           & Avg$_{F1}$ & Avg$_{F1}$                             & Avg$_{Sim}$                            \\ \midrule
0.4                        & 40.9       & 36.2                                   & 20.9                                   \\
0.5                        & 41.7       & 37.1                                   & 21.3                                   \\
0.6                        & 42.3       & 37.5                                   & 21.8                                   \\
0.7                        & 42.1       & 35.6                                   & 21.2                                   \\
0.8                        & 39.7       & 31.5                                   & 19.9                                   \\ \bottomrule
\end{tabular}
\vspace{-1mm}
\caption{Threshold analysis on E.T. Bench data.}
\label{tab:threshold}
\end{table}

\section{More Experimental Results}
\subsection{Performance on E.T. Bench Complex Dataset}
Here we also compare the performance on E.T. Bench Complex dataset~\citesupp{etbench_nips24_supp} that involves two sub-tasks: temporal event matching and grounded video question answering. The results are shown in Tab.~\ref{tab: complex}. It can be seen that our approach also outperforms the existing state of the art by large margins, further demonstrating its superiority.

\subsection{Extension to General QA Tasks}

We test our model on general video question answering benchmarks (MVBench~\citesupp{mvbench_supp} and Video-MME~\citesupp{video-mme_supp}). To enhance basic instruction-following capability, we incorporate automatically constructed multiple-choice questions during the proposed factorized preference optimization process. Due to our proposed factorized preference data synthesis, we can easily generate diverse distractor options based on different causes of failure and combine them with the original correct answer to form multiple-choice questions, without requiring additional external data sources.

As shown in Tab.~\ref{tab:general_video}, our method outperforms the grounding-focused counterpart E.T. Chat~\citesupp{etbench_nips24_supp} and achieves results comparable to some general video understanding models (e.g., Video-LLaVA~\citesupp{video_llava_emnlp24_supp}) trained on large-scale generic data, but are usually less effective on grounding. We attribute the performance gap between our model and recent SOTA methods~\citesupp{qwen2.5vl_supp, livecc_cvpr25_supp} to the absence of large-scale generic pretraining and the relatively smaller model size. Incorporating such data and scaling up the model could further improve our framework. Meanwhile, it is also worth exploring how to train a model that can simultaneously achieve strong general reasoning and accurate temporal grounding.

\subsection{Cost of Frame-Wise Similarity Calculation}

Since the designed \texttt{<evi>} token involves additional frame-wise feature similarity computation for temporal grounding and visual semantic aggregation beyond the standard autoregressive decoder, it is natural to evaluate the associated computational cost. Such a frame-wise similarity calculation process is actually lightweight, taking less than 0.4 ms per token generation on a single 3090 GPU—only 1.4\% of the total network forwarding time (29 ms).

\subsection{Sensitivity Analysis on Similarity Threshold} \label{exp:threshold}

As shown in Tab.~\ref{tab:threshold}, performance is relatively robust across different threshold values for salient frame identification during inference, and the intuitive choice of 0.5 already yields acceptable results. Overall, an overly high threshold causes information loss, while an overly low one introduces less relevant context. The best performance is achieved at a threshold of 0.6.

\section{More about the Factorized Data Synthesis}
\begin{figure*}[htbp]
\centering
\includegraphics[width=0.77\textwidth]{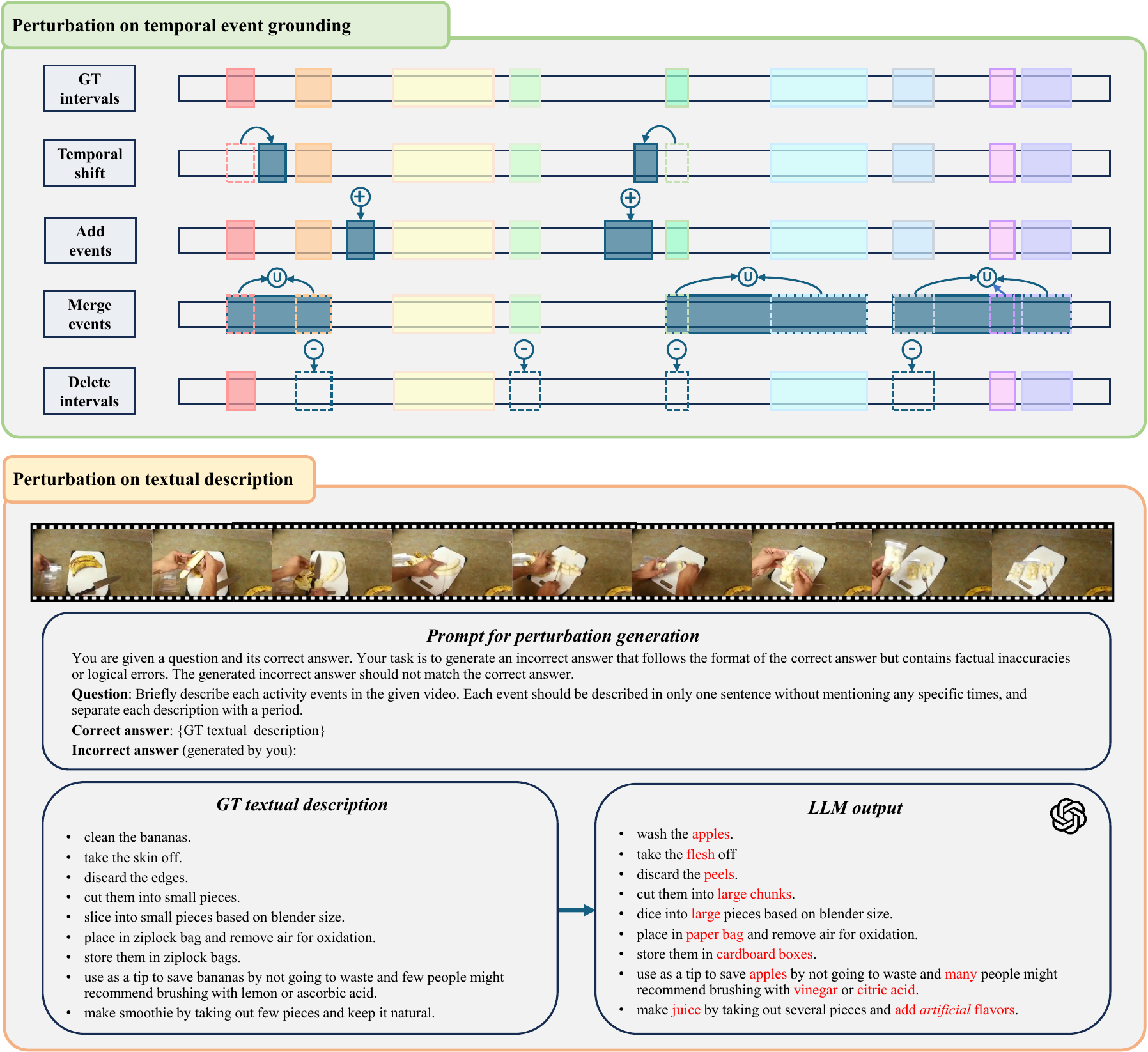}
\caption{An illustrative example of the data synthesis approach.}
\label{fig:data_gen}
\end{figure*}

As mentioned in the main paper, we mainly focus on two main factors: temporal event grounding and textual response, where each factor can be further categorized into multiple sub-factors. For temporal event grounding aspects, sub-factors include temporal localization shift, randomly adding or deleting grounded events (corresponding to the simulation of false positives and missed detection), and merging multiple events into one (corresponding to the simulation of a lack of fine-grained distinction in event boundaries). A full demonstration example can be found at Fig.~\ref{fig:data_gen}. For textual response aspect, this type of perturbation modifies the semantic correctness of the textual response. It includes sub-types such as distorting key information, which disrupts critical content, and repeating responses, a common failure mode observed in video LLMs. Except for the repeating factor, we prompt an off-the-shelf LLM~\citesupp{qwen2.5_supp} to generate a distracted response based on the original correct event-level response.

\section{Visualization Results}
Here we provide qualitative results to better demonstrate the capability of our approach. Based on the input-output format, we categorize different tasks into three main types: (1) Dense captioning related task, which requires grounding multiple events throughout the entire video while also providing a textual description for each grounded event. (2) Grounding-focused task (e.g., temporal video grounding, action localization, etc.), which includes single-event grounding and multi-event grounding.  (3) Temporally grounded video question answering, which involves answering the user's open-ended questions while also providing the temporal position of the answer (evidence). We also visualize the prediction result from the recent SOTA method~\citesupp{etbench_nips24_supp} for comparison. 
Note that in the response from D$^2$VLM, all temporal information is derived from the generated \texttt{<evi>} token through the conversion process illustrated in the main paper. For each input, D$^2$VLM will first perform pure evidence grounding, followed by interleaved text-evidence generation (here we denote this as its actual response part). 
We show the converted time-involved text for both stages, where the actual response stage begins after the ``Answer:" marker.

\begin{figure*}[t]
\centering
\includegraphics[width=0.77\textwidth]{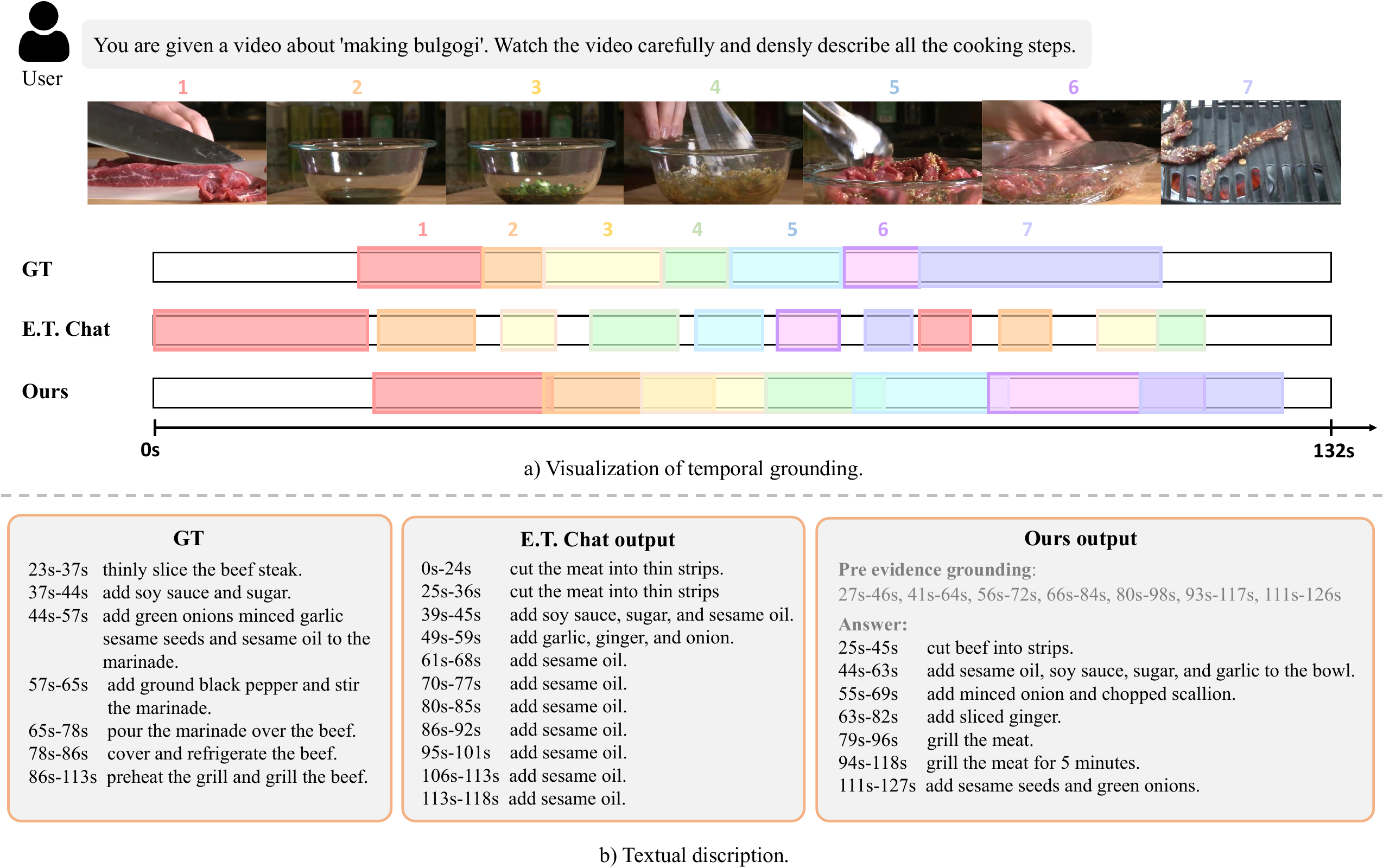}
\caption{A qualitative example for dense captioning task.}
\label{fig:dvc}
\end{figure*}

\noindent \textbf{Dense captioning task.} From Fig.~\ref{fig:dvc}, we can observe that: (1) Compared with the recent counterpart, our method can better localize the individual events. (2) Our method also generates more coherent and meaningful textual description, whereas the compared method often fails to do so and repeatedly generates similar content. These results essentially demonstrate the superiority of our approach in both event grounding and textual generation.

\begin{figure*}[t]
\centering
\includegraphics[width=0.8\textwidth]{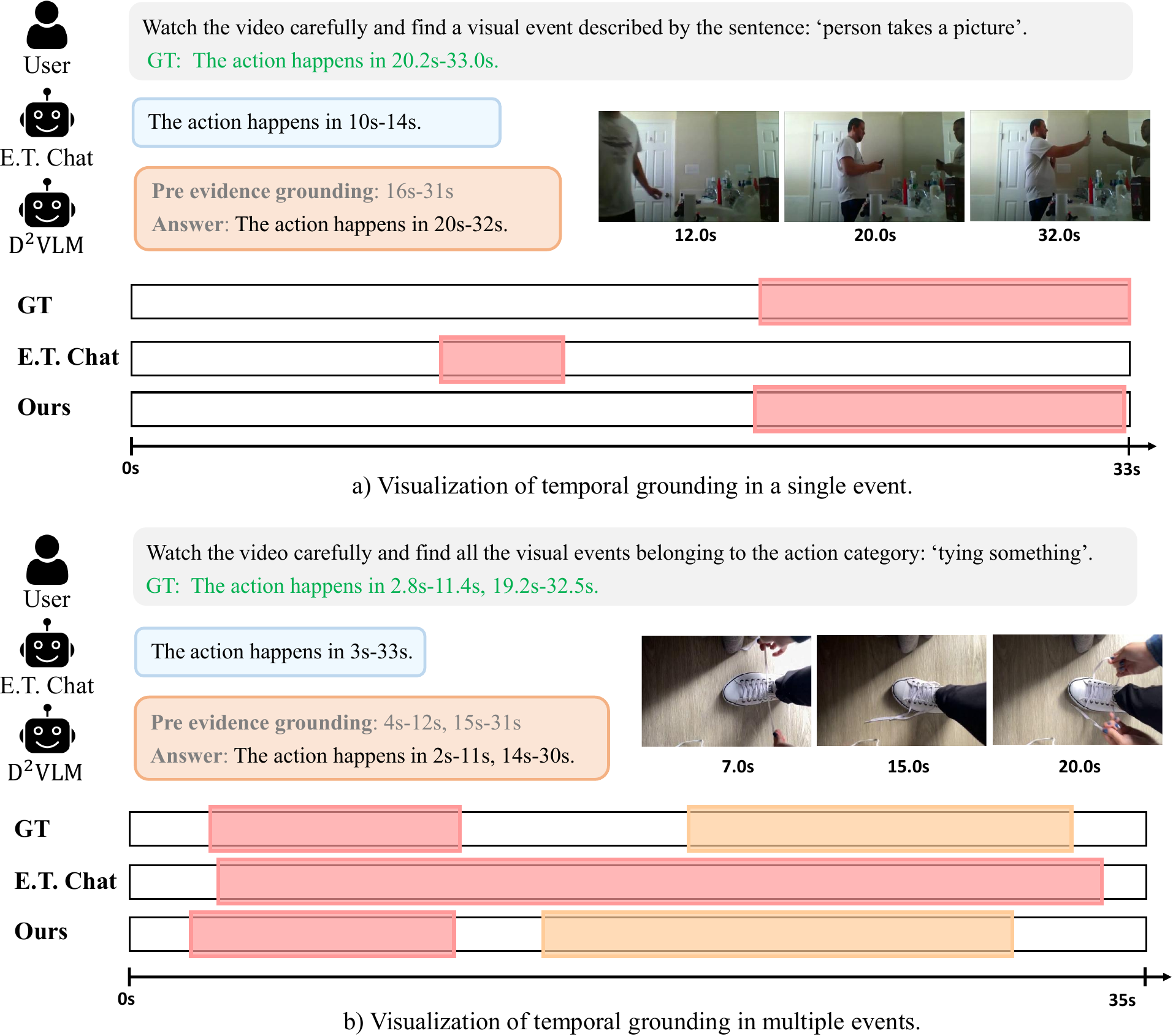}
\caption{Qualitative examples for grounding-focused task.}
\label{fig:tal}
\end{figure*}

\noindent \textbf{Grounding-focused task.} The qualitative examples are shown in Fig.~\ref{fig:tal}. It can be observed that: (1) Compared with recent SOTA method~\citesupp{etbench_nips24_supp}, our approach can localize the desired temporal event position more accurately (Fig.~\ref{fig:tal} (a)). (2) Our method also better distinguishes the boundaries between individual events, demonstrating its fine-grained discrimination capability (Fig.~\ref{fig:tal} (b)).

\begin{figure*}[t]
\centering
\includegraphics[width=0.8\textwidth]{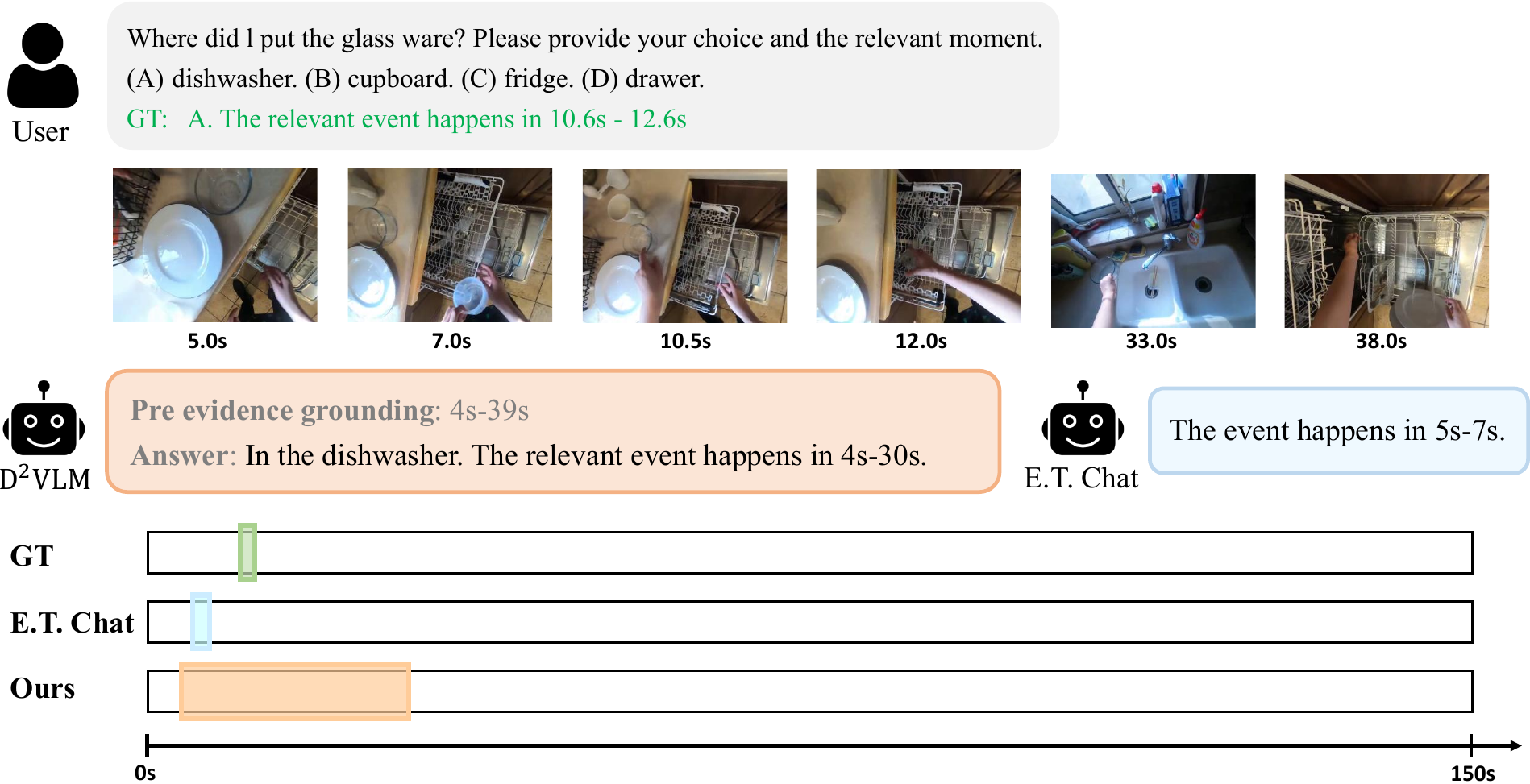}
\caption{A qualitative example for temporally grounded video question answering task.}
\label{fig:gvq}
\end{figure*}

\noindent \textbf{Temporally grounded video question answering.} The example is shown in Fig.~\ref{fig:gvq}. It can be observed that: (1) Our method can correctly answer the question, while the compared method fails to follow the instruction given by the user (i.e., only responding to the temporal evidence position without answering the question). (2) Our method can also provide more reliable temporal evidence grounding.

\makeatletter
\@ifundefined{c@NAT@ctr@supp}{}{
  \setcounter{NAT@ctr@supp}{0}}
\setcounter{NAT@ctr}{0}
\makeatother

{
    \small
    \bibliographystylesupp{ieeenat_fullname}
    \bibliographysupp{supp}
}

\end{document}

%% file: ICCV2025-Author-Kit-Feb/tab/quan.tex
\begin{table*}[t]
\setlength{\tabcolsep}{3pt}
\resizebox{\linewidth}{!}{
\begin{tabular}{lccccccccccccc}
\toprule
\multirow{2}{*}{Method} & \multirow{2}{*}{Year} & \multicolumn{6}{c}{E.T. Bench Grounding}                                                      & \multicolumn{6}{c}{E.T. Bench Dense Captioning}                                               \\
\cmidrule(lr){3-8} \cmidrule(lr){9-14}
                        &                       & TVG$_{F1}$           & EPM$_{F1}$           & TAL$_{F1}$           & EVS$_{F1}$           & VHD$_{F1}$           & Avg$_{F1}$        & DVC$_{F1}$       & DVC$_{Sim}$      & SLC$_{F1}$       & SLC$_{Sim}$      & Avg$_{F1}$       & Avg$_{Sim}$      \\ \midrule

Video-ChatGPT-7B~\cite{video_bench_arxiv23}        & ACL'24              & 7.0           & 1.3           & 15.1          & 8.4           & 28.8          & 12.1          & 8.8           & 11.3          & 5.7           & 10.2          & 7.3           & 10.8          \\
Video-LLaVA-7B~\cite{video_llava_emnlp24}          & EMNLP'24            & 7.0           & 1.9           & 15.0          & 0.3           & 28.9          & 10.6          & 28.0          & 15.0          & 0.9           & 8.3           & 14.5          & 11.7          \\
LLaMA-VID-7B~\cite{llama_vid}            & ECCV'24             & 5.5           & 1.2           & 8.0           & 1.4           & 30.0          & 9.2           & 27.1          & 12.6          & 5.2           & 11.1          & 16.2          & 11.9          \\
Video-LLaMA-2-7B~\cite{video_llama_2}        & arXiv'24              & 0.1           & 0.0           & 0.0          & 0.0           & 1.5          & 0.3          & 0.6           & 14.5          & 0.0           & 15.2          & 0.3           & 14.9          \\
PLLaVA~\cite{pllava}        & arXiv'24              & 6.9           & 1.1           & 5.7          & 0.3           & 28.9          & 8.6          & 13.3           & 10.6          & 9.7           & 11.8          & 11.5           & 11.2          \\
VTimeLLM-7B~\cite{vtimellm_cvpr24}             & CVPR'24             & 7.6           & 1.9           & 18.2          & 15.9          & 28.9          & 14.5          & 12.4          & 13.1          & 8.7           & 6.4           & 10.6          & 9.8           \\
TimeChat-7B~\cite{timechat_cvpr24}             & CVPR'24             & 26.2          & 3.9           & 10.1          & 29.1          & 40.5          & 22.0          & 16.6          & 12.5          & 5.6           & 9.2           & 11.1          & 10.9          \\
LITA-13B~\cite{lita_eccv24}                & ECCV'24             & 22.2          & 4.6           & 18.0          & {29.7}    & 23.9          & 19.7          & 39.7          & 17.2          & 21.0          & 12.2          & 30.4          & 14.7          \\
VTG-LLM-7B~\cite{vtgllm_aaai25}              & AAAI'25             & 15.9          & 3.7           & 14.4          & 26.8          & 48.2          & 21.8          & {40.2}    & 18.6          & 20.8          & 14.4          & 30.5          & 16.5          \\
Qwen2.5-VL-7B*~\cite{qwen2.5vl}              & arXiv'25             & 46.6          & 9.3           & 32.2          & 19.9          & \textbf{68.6}          & 35.3          & 39.0    & 21.4          & 25.5          & 15.1          & 32.2          & 18.2          \\

E.T.Chat-3.8B~\cite{etbench_nips24}           & NeurIPS'24          & {38.6}    & {10.2}    & {30.8}    & 25.4          & {62.5}    & {33.5}    & 38.4          & {19.7}    & {24.4}    & {14.6}    & {31.4}    & {17.2}    \\

\rowcolor[HTML]{ECF4FF}
D$^2$VLM-3.8B (Ours)              &  ICCV'25                  &  \textbf{60.2} & \textbf{14.4} & \textbf{33.4} & \textbf{35.2} & 68.2 & \textbf{42.3} & \textbf{48.3} & \textbf{25.5} & \textbf{26.7} & \textbf{18.1} & \textbf{37.5} & \textbf{21.8} \\ \bottomrule
\end{tabular}
}
\vspace{-2mm}
\caption{Performance comparison on E.T. Bench-Grounding dataset (5 sub tasks: TVG: Temporal Video Grounding; EPM: Episodic Memory; TAL: Temporal Action Localiztion; EVS: Extractive Video Summarization; VHD: Video Highlight Detection) and E.T. Bench-Dense Captioning dataset (2 sub tasks: DVC: Dense Video Captioning; SLC: Step Localization and Captioning). *: Implemented by us, with more details in the supplementary material.}
\label{tab:quan}
\end{table*}

\begin{table}[t]
\footnotesize
\resizebox{\linewidth}{!}{
\begin{tabular}{lccc}
\toprule
Method         & Year                 & R@1(IoU=0.5)   & R@1(IoU=0.7)   \\ \midrule
TimeChat-7B~\cite{timechat_cvpr24}    & CVPR'24            & 32.2          & 13.4          \\
VTimeLLM-7B~\cite{vtimellm_cvpr24}    & CVPR'24            & 27.5          & 11.4          \\
VTimeLLM-13B~\cite{vtimellm_cvpr24}   & CVPR'24            & 34.3          & 14.7          \\
Momentor-7B~\cite{momentor_icml2024}    & ICML'24            & 26.6          & 11.6          \\
HawkEye-7B~\cite{hawkeye_arxiv24}     & arXiv'24           & 31.4          & 14.5          \\
VTG-LLM-7B~\cite{vtgllm_aaai25}     & AAAI'25            & 33.8          & 15.7          \\
TRACE-7B~\cite{trace_iclr25}       & ICLR'25            & 40.3          & 19.4          \\
VideoChat-T-7B~\cite{timesuite_iclr25} & ICLR'25            & 48.7          & 24.0          \\
E.T.Chat-3.8B~\cite{etbench_nips24}  & NeurIPS'24         & 45.9          & 20.0          \\
\rowcolor[HTML]{ECF4FF}
D$^2$VLM-3.8B (Ours) & ICCV'25 & \textbf{50.3} & \textbf{26.0} \\ \bottomrule

\end{tabular}}
\vspace{-2mm}
\caption{Performance comparison on Charades-STA.}
\label{tab: charades}
\end{table}

\begin{table}[t]
\footnotesize
\resizebox{\linewidth}{!}{
\begin{tabular}{lcccc}
\toprule
Method         & Year      & F1            & CIDEr         & SODA\_c      \\ \midrule
TimeChat-7B~\cite{timechat_cvpr24}    & CVPR'24 & 12.6          & 3.4           & 1.2          \\
VTG-LLM-7B~\cite{vtgllm_aaai25}     & AAAI'25 & 17.5          & 5.0           & 1.5          \\
TRACE-7B~\cite{trace_iclr25}       & ICLR'25 & 22.4          & 8.1           & 2.2          \\
\rowcolor[HTML]{ECF4FF}
D$^2$VLM-3.8B (Ours) & ICCV'25      & \textbf{26.4} & \textbf{10.6} & \textbf{3.2} \\ \bottomrule
\end{tabular}}
\vspace{-2mm}
\caption{Performance comparison on YouCook2.}
\label{tab: youcook2}
\end{table}

%% file: ICCV2025-Author-Kit-Feb/tab/ab1.tex
\begin{table*}[t]
\footnotesize
  \begin{minipage}{0.62\linewidth}
    \centering
    \resizebox{1\linewidth}{!}{
\begin{tabular}{ccc|c|cc}
\toprule
\multirow{2}{*}{\begin{tabular}[c]{@{}c@{}}Decomposed\\ Objective\end{tabular}} & \multirow{2}{*}{\begin{tabular}[c]{@{}c@{}}Interleaved Text-evi\\ Generation\end{tabular}} & \multirow{2}{*}{\begin{tabular}[c]{@{}c@{}}Consistency \\ Constraint\end{tabular}} & Grounding & \multicolumn{2}{c}{Dense Captioning} \\ 
                                                                                &                                                                                            &                                                                                    & Avg$_{F1}$   & Avg$_{F1}$          & Avg$_{Sim}$          \\ \midrule
                                                                                &                                                                                            &                                                                                    & 21.2     & 14.3             & 11.3             \\
\checkmark                                                      &                                                                                            &                                                                                    & 28.9     & 23.1            & 16.0             \\
\checkmark                                                      & \checkmark                                                                 &                                                                                    & 35.6     & 34.3            & 19.8             \\
\checkmark                                                      & \checkmark                                                                 & \checkmark                                                          & \textbf{39.5} & \textbf{35.0}    & \textbf{21.2}    \\ \bottomrule
\end{tabular}
}
    \vspace{-2mm}
    \caption{Component effect of the designed generation objective.}
    \label{tab:ab1}
\end{minipage}
\hspace{10pt}
\begin{minipage}{0.3\linewidth}
    \centering
    \renewcommand{\arraystretch}{1.55} 
    \resizebox{1\linewidth}{!}{
\begin{tabular}{c|c|cc}
\toprule
     \multirow{2}{*}{FPO}            & Grounding & \multicolumn{2}{c}{Dense Captioning} \\
                 & Avg$_{F1}$   & Avg$_{F1}$          & Avg$_{Sim}$          \\ \midrule
$\times$           & 39.5     & 35.0            & 21.2              \\
\checkmark         & \textbf{42.3}      & \textbf{37.5}             & \textbf{21.8}              \\ \bottomrule
\end{tabular}
}   
    \vspace{-2mm}
    \caption{Effect of the proposed FPO.}
    \label{tab:ab3}
\end{minipage}
\end{table*}

%% file: ICCV2025-Author-Kit-Feb/tab/ab2.tex
\begin{table}[t]
\centering
\footnotesize
    \setlength{\tabcolsep}{2pt}
    \resizebox{0.9\linewidth}{!}{
\begin{tabular}{l|c|cc}
\toprule
    \multirow{2}{*}{Method}                        & Grounding & \multicolumn{2}{c}{Dense Captioning} \\
                            & Avg$_{F1}$   & Avg$_{F1}$          & Avg$_{Sim}$          \\ \midrule
w/o event-level modeling    & 26.1     & 33.4             & 16.2             \\
w/o visual semantic capture & 37.1     & 27.5            & 17.7              \\
Full design                  & \textbf{39.5}     & \textbf{35.0}            & \textbf{21.2}              \\ \bottomrule
\end{tabular}
}   
    \vspace{-2mm}
    \caption{Component effect within evidence token design.}
    \label{tab:ab2}
\end{table}